\documentclass[sigconf,nonacm]{acmart}

\usepackage{graphicx}
\usepackage{balance}
\usepackage{tabularx}
\usepackage{xcolor}
\usepackage{booktabs}
\usepackage{colortbl}

\AtBeginDocument{%
  }

\acmConference[]{}{}{}
\acmBooktitle{}
\acmISBN{}
\acmDOI{}

\begin{document}

\title{STEP: Stepwise Curriculum Learning for Context-Knowledge Fusion in Conversational Recommendation}

\author{Zhenye Yang}
\affiliation{%
  \institution{School of Computer Science (National Pilot Software Engineering School)\\
    Beijing University of Posts and Telecommunications}
  \city{Beijing}
  \country{China}
}
\email{yangzhenye@bupt.edu.cn}
\authornotemark[2]

\author{Jinpeng Chen}
\affiliation{%
  \institution{School of Computer Science (National Pilot Software Engineering School)\\
    Beijing University of Posts and Telecommunications}
  \city{Beijing}
  \country{China}
}
\email{jpchen@bupt.edu.cn}
\authornote{{Corresponding author.}}
\authornote{Also with Key Laboratory of Trustworthy Distributed Computing and Service (BUPT), Ministry of Education.}

\author{Huan Li}
\affiliation{%
  \institution{The State Key Laboratory of Blockchain and Data Security\\
    Zhejiang University}
  \city{Hangzhou}
  \country{China}
 }
\email{lihuan.cs@zju.edu.cn}

\author{Xiongnan Jin}
\affiliation{%
  \institution{School of Artificial Intelligence\\ Shenzhen University}
  \city{Shenzhen}
  \country{China}}
\email{xiongnanjin@szu.edu.cn}
\authornote{Also with National Engineering Laboratory for Big Data System Computing Technology, Shenzhen University.}

\author{Xuanyang Li}
\affiliation{%
  \institution{Beijing University of Posts and Telecommunications}
  \city{Beijing}
  \country{China}
}
\email{lixuanyang@bupt.edu.cn}

\author{Junwei Zhang}
\affiliation{%
  \institution{Beijing University of Posts and Telecommunications}
  \city{Beijing}
  \country{China}
}
\email{buptscszjw@bupt.edu.cn}

\author{Hongbo Gao}
\affiliation{%
  \institution{USTC}
  \city{Hefei}
  \country{China}
  }
\email{ghb48@ustc.edu.cn}

\author{Kaimin Wei}
\affiliation{%
  \institution{Jinan University}
  \city{Guangzhou}
  \country{China}}
\email{kaiminwei@jnu.edu.cn}

\author{Senzhang Wang}
\affiliation{%
  \institution{Central South University}
  \city{Changsha}
  \country{China}}
\email{szwang@csu.edu.cn}

\renewcommand{\shortauthors}{Zhenye Yang et al.}

\begin{abstract}
Conversational recommender systems (CRSs) aim to proactively capture user preferences through natural language dialogue and recommend high-quality items. To achieve this, CRS gathers user preferences via a dialog module and builds user profiles through a recommendation module to generate appropriate recommendations. However, existing CRS faces challenges in capturing the deep semantics of user preferences and dialogue context. In particular, the efficient integration of external knowledge graph (KG) information into dialogue generation and recommendation remains a pressing issue. Traditional approaches typically combine KG information directly with dialogue content, which often struggles with complex semantic relationships, resulting in recommendations that may not align with user expectations.

To address these challenges, we introduce STEP, a conversational recommender centered on pre-trained language models that combines curriculum-guided context-knowledge fusion with lightweight task-specific prompt tuning. At its heart, an F-Former progressively aligns the dialogue context with knowledge-graph entities through a three-stage curriculum, thus resolving fine-grained semantic mismatches. The fused representation is then injected into the frozen language model via two minimal yet adaptive prefix prompts: a conversation prefix that steers response generation toward user intent and a recommendation prefix that biases item ranking toward knowledge-consistent candidates. This dual-prompt scheme allows the model to share cross-task semantics while respecting the distinct objectives of dialogue and recommendation. Experimental results show that STEP outperforms mainstream methods in the precision of recommendation and dialogue quality in two public datasets. Our code is available: \textit{https://github.com/Alex-bupt/STEP}.
\end{abstract}



\maketitle

\begin{figure}[htbp]
\centerline{\includegraphics[width=0.95\linewidth]{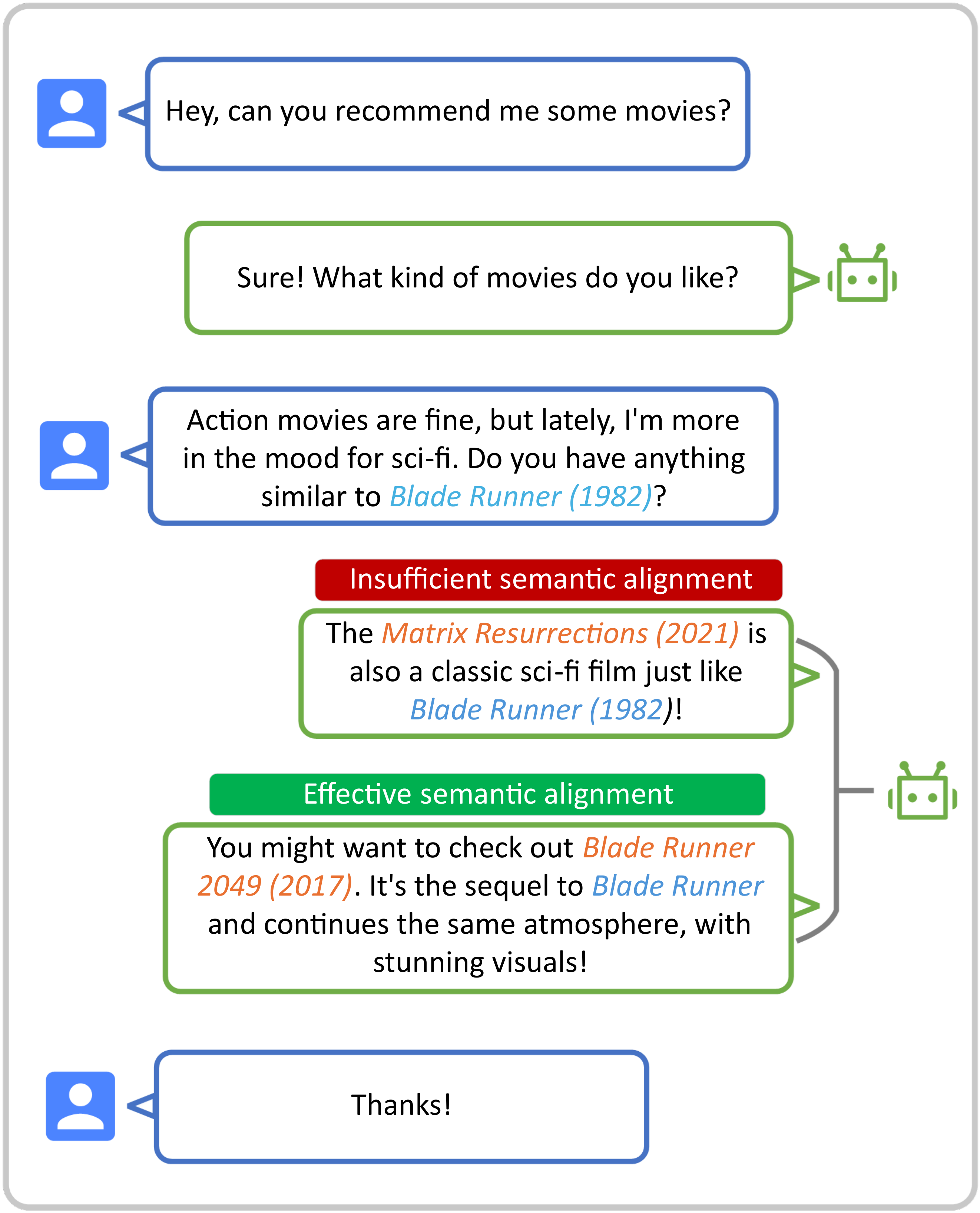}}
\caption{An example of a user requesting a conversational recommendation system for movie recommendations.}
\Description{Example dialogue illustrating the difference between insufficient and effective semantic alignment in a conversational recommender. The user requests a science fiction movie similar to Blade Runner (1982). In the insufficient alignment path, the system recommends The Matrix Resurrections (2021). In the effective alignment path, the system identifies the sequel relationship and recommends Blade Runner 2049 (2017). Text annotations label the two paths as “Insufficient semantic alignment” and “Effective semantic alignment” to highlight the contrast in retrieval and fusion strategies.}
\label{fig:conversation}
\end{figure}

\section{Introduction}

With the rapid development of recommender systems, conversational recommendation systems have become a research hotspot, offering personalized recommendations via natural language interactions~\cite{christakopoulou2016towards,li2018towards}. Through dialogue, CRS collects user preferences and intentions, enhancing both user engagement and system interactivity.

Conversational recommender systems (CRSs) consist of two interdependent modules: dialogue understanding, which processes user utterances to infer needs, preferences and contextual nuances; and recommendation inference, which generates or ranks items based on the inferred user state. To infuse structured knowledge into both modules, many approaches leverage external knowledge graphs (KGs) that encode item attributes and semantic relations~\cite{chen2019towards,dao2024broadening,wang2022towards,zhang2023variational,xu2024enhancing,zhou2020improving}. However, naively fusing KG embeddings with rich dialogue representations often leaves a semantic gap and can even degrade recommendation relevance. Recent solutions have begun to address this: KGSF maximizes mutual information to better align dialogue and KG semantics~\cite{zhou2020improving}, VRICR employs variational inference to mitigate KG incompleteness~\cite{zhang2023variational}, and DCRS introduces knowledge-aware contrastive learning to sharpen entity representations~\cite{dao2024broadening}. Yet, despite these advances, pre-trained language models still struggle to leverage dialogue context for retrieving and integrating the most pertinent KG information, limiting their ability to capture nuanced user intents.

To illustrate this challenge of semantic alignment described above, Figure~\ref{fig:conversation} presents a dialogue in which the user requests a science‐fiction film akin to \textit{Blade Runner}. An effective recommender must not only recognize \textit{Blade Runner} as Ridley Scott's dark and philosophically probing sci-fi thriller, but also leverage the knowledge graph to identify its official sequel, \textit{Blade Runner 2049}. If semantic alignment fails and the system cannot draw on KG information, it may instead surface other popular sci‐fi titles (e.g., \textit{The Matrix Resurrections}), thereby overlooking the user’s implicit desire for the sequel.

Pre-trained language models (PLMs) have been embraced as a unified backbone for CRS, enabling end-to-end optimization of both dialogue generation and item recommendation~\cite{brown2020language,gao2020making,xu2024sequence}. PLM-based frameworks such as RID, which fine-tunes large PLM alongside a pre-trained R-GCN to inject structural KG embeddings during generation~\cite{wang2021finetuning}; UniCRS, which semantically fuses dialogue and KG representations via knowledge-enhanced prompt learning~\cite{wang2022towards}; and DCRS, which employs knowledge-aware contrastive retrieval to prepend in-context demonstrations as soft prompts have all improved alignment between dialogue context and external knowledge~\cite{dao2024broadening}. However, relying on relatively simple fusion modules, such as single-layer concatenation or basic cross-attention to merge dialogue context and KG information, these approaches fall short of empowering PLM to harness the knowledge graph’s valuable information, which in turn undermines their ability to discern subtle user intentions.

To address these challenges, we reconceptualize context knowledge fusion as a curriculum of alignment objectives and introduce STEP, a framework that progressively bridges the semantic gap between dialogue and knowledge graphs. Rather than statically combining modalities, STEP employs a three-stage curriculum for semantic alignment~\cite{wang2021survey}, fine-grained discrimination through hard negative triplet refinement, and nuanced consolidation with auxiliary matching to adaptively calibrate representations across heterogeneous knowledge spaces. We embed this curriculum in the F-Former module, which uses learnable cross-modal queries and a coarse-to-fine scheduling strategy to drive contextual fusion. Crucially, STEP also adopts a dynamic prompt adaptation mechanism that injects these fused context–knowledge embeddings into the PLM's prompt space~\cite{brown2020language,jannach2021survey,wang2021finetuning}, ensuring that both dialogue generation and item recommendation are directly informed by integrated semantics. By treating fusion as an evolving process rather than a fixed engineering pipeline, STEP optimizes adaptive alignment with evolving graph information and faithful capture of user intents.

Our main contributions are as follows.
\begin{itemize}
    \item We propose STEP, a conversational recommendation system that integrates a curriculum-guided F-Former architecture with efficient prompt learning to jointly optimize dialogue generation and item recommendation.
    \item We design the F-Former module to include three subtasks and employ dynamic weight scheduling to realize a “from easy to hard” curriculum learning strategy that progressively enhances the fusion of dialogue context and knowledge graph semantics.
    \item Extensive experiments on multiple public datasets demonstrate that STEP surpasses existing state-of-the-art methods in both recommendation accuracy and dialogue quality.
\end{itemize}

\section{Related Works}

\subsection{Conversational Recommendation}

Conversational Recommender Systems aim to capture user preferences and deliver relevant recommendations through multi-turn dialogues. Current approaches fall into two categories: predefined operation-based and generative CRS~\cite{christakopoulou2016towards,gao2021advances,jannach2021survey}. Predefined operation-based CRS relies on fixed interaction patterns (e.g., slot filling, attribute selection) to reduce interaction rounds, often using reinforcement learning\cite{jannach2021survey} or multi-armed bandits~\cite{christakopoulou2016towards,xie2021comparison}. However, their dependence on templates limits their adaptability to complex scenarios. Generative CRS focuses on separating conversation and recommendation tasks into independent modules. While they improve natural dialogue generation and preference capturing, they struggle with semantic inconsistency between modules, particularly in multi-turn, multi-item dialogues. Solutions such as shared knowledge resources (e.g., knowledge graphs)~\cite{chen2019towards,DBLP:conf/acl/LuBSMCWH21} or semantic alignment strategies~\cite{zhou2020improving,wang2022towards} have made progress but still face challenges in maintaining recommendation relevance.

The emergence of PLM has further advanced CRS research~\cite{wang2021finetuning,wang2022towards}. While early methods froze PLM parameters~\cite{petruzzelli2024towards,wang2022towards}, limiting their utility in complex scenarios, recent studies employ prompt learning to unify semantic representations for both recommendation and dialogue tasks~\cite{lester2021power,dao2024broadening,wang2022towards}.

\subsection{Semantic fusion in recommendation}

Semantic fusion refers to the integration of information from diverse sources or modalities to create a comprehensive representation of user preferences. It plays a key role in areas such as personalized, content-based, and social recommendations~\cite{yuan2022semantic,xu2024sequence,meng2020heterogeneous,zheng2024adapting}. Early methods treated each information source independently and used basic fusion techniques like vector concatenation, weighted averaging, or shared representations. For instance, in multimodal systems, user behavior data and social network information might be processed separately and then combined. However, these approaches often fail to capture the deep semantic relationships within complex and high-dimensional data, resulting in suboptimal performance.

To address these limitations, recent research has adopted attention mechanisms~\cite{lin2023cola,deng2024task} and deep neural networks~\cite{al2024improved} to strengthen semantic coherence across data sources. A notable example is Q-Former~\cite{li2023blip}, a query-based Transformer that bridges visual encoders and language models by using learnable query vectors to interact with visual features and extract key semantic information. Q-Former achieves deep semantic fusion across image and text modalities, improving the model’s ability to interpret cross-modal relationships.

Building on the Q-Former approach, we adapted and extended its principles to develop the F-Former framework. This framework integrates multi-source information from both contextual data and knowledge graphs, facilitating more effective semantic fusion. As a result, it enhances both recommendation and conversation tasks, addressing challenges in multi-source and multi-turn interaction scenarios.

\section{Problem Definition}
To build an effective CRS, we formally define three key tasks. Let $u$ represent the user, $\textit{i} \in I$ an item, and $w \in V$ a word in the vocabulary. The system aims to dynamically capture user preferences through natural language interactions and recommend items that align with user needs.

\textbf{Definition 1 (Dialogue Representation):}  
A dialogue is \(C = \{s_1, \dots, s_t\}\), where each utterance \(s_i = \{w_1, \dots, w_n\}\) is drawn from vocabulary \(V\).  A compact representation captures semantic and contextual dependencies across turns, enabling the system to track evolving user preferences.

\textbf{Definition 2 (Knowledge Graph Modeling):}  
A knowledge graph is \(G = \langle E, R, \mathcal{T}\rangle\), where \(E\) is the set of entities, \(R\) the set of relations, and \(\mathcal{T} \subseteq E \times R \times E\) the triples \((e_h, r, e_t)\).  Embedding items as entities in \(G\) allows the model to leverage external knowledge for richer, more personalized recommendations.  

\textbf{Definition 3 (Recommendation Tasks):} At turn $t$, given dialogue history $C = \{s_1, s_2, \dots, s_t\}$ and item set $I$, the system must:

\begin{enumerate}
    \item \textbf{Item Recommendation Task}: Select a subset \(I_s \subset I\) that meets user needs. If no items are required in a turn, then \(I_s = \emptyset\).
    \item \textbf{Dialogue Generation Task}: Generate a response \(R_t = \{w_1, w_2, \dots, w_m\}\) containing items in \(I_s\) to continue the conversation.
\end{enumerate}

\section{Approach}
STEP is a knowledge-enhanced conversational recommender built on a pretrained LM: it encodes item–entity interactions via graph relation convolution (Sec.~4.1), aligns them with dialogue through the F-Former (Sec.~4.2) to enrich item embeddings, turns those into prompts for response generation (Sec.~4.3), and finally yields personalized suggestions in the recommendation module (Sec.~4.4; Fig.~\ref{fig:model}).

\subsection{Graph Relation Convolution}

The PLM employed in STEP is based on DialoGPT~\cite{DBLP:conf/acl/ZhangSGCBGGLD20}. To enrich item representations, we incorporate the external knowledge graph DBpedia~\cite{auer2007dbpedia,chen2019towards} and employ a Relational Graph Convolutional Network (RGCN) to perform graph convolutions for learning node embeddings. The convolutional process of RGCN is defined as:

\begin{figure*}[ht]
    \centering
    \includegraphics[width= 0.98\textwidth]{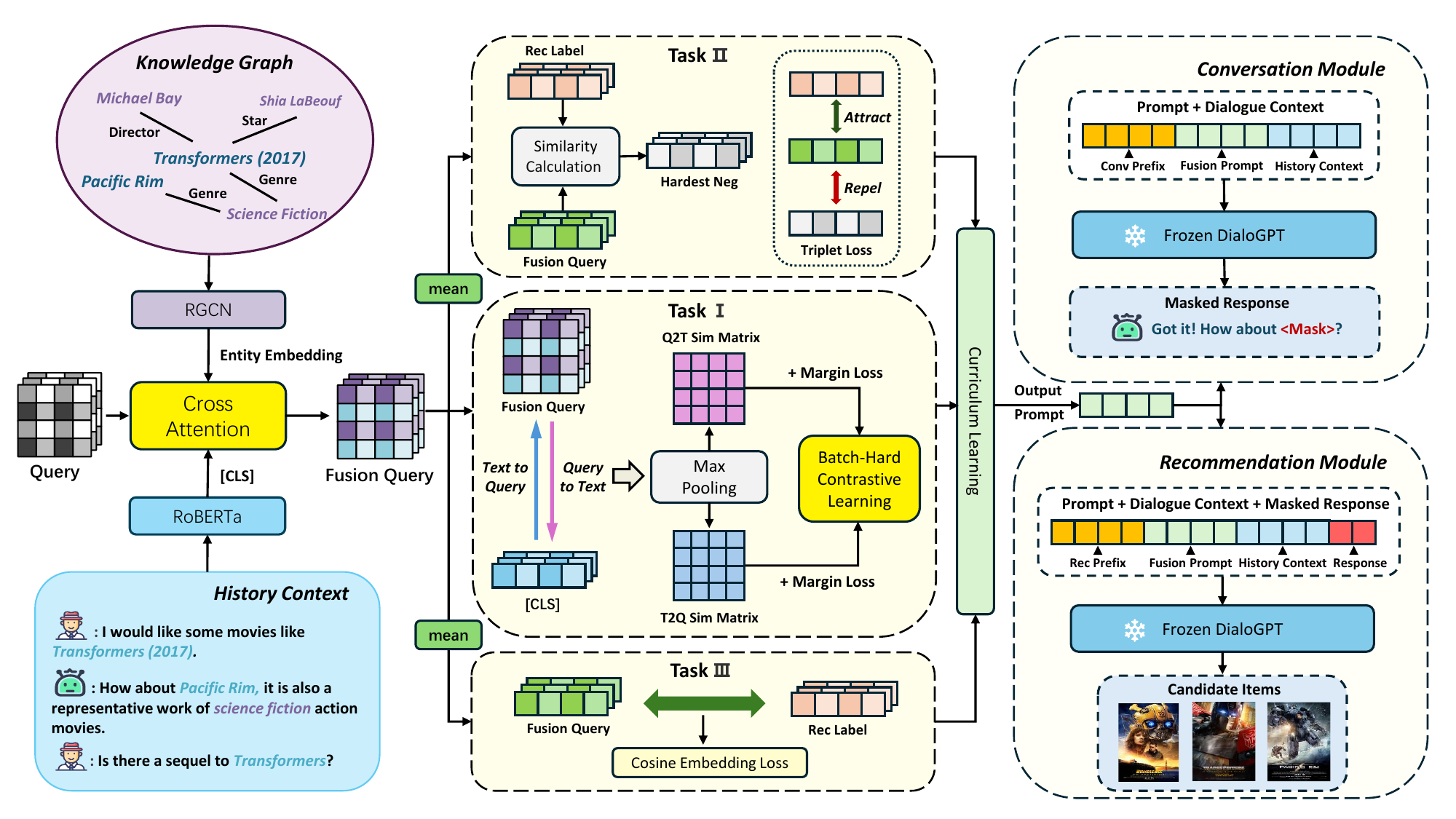}
    \caption{The architecture of the STEP model proposed in this paper includes three main modules: the F-Former knowledge-context fusion module based on curriculum learning, the response generation module based on prompt learning, and the item recommendation module.}
    \Description{Overall architecture and training flow of the STEP framework. The top section shows parallel encoding of a knowledge graph and dialogue context: entities are encoded via an R-GCN, and learnable query vectors attend first to entity embeddings and then to the text [CLS] token through cross-modal attention, producing fused queries. These are used to compute query-to-text and text-to-query similarity matrices for batch-hard contrastive learning with a margin, along with a triplet loss aligning queries to recommendation labels, and an auxiliary cosine embedding loss, corresponding to Task I, Task II, and Task III. A curriculum learning arrow indicates the progressive introduction of these tasks from easy to hard. The bottom section shows the response generation and item recommendation modules: the fused representation is injected into a conversational prefix and a recommendation prefix, which are fed into a frozen DialoGPT to generate responses and rank candidate items. In generation, placeholders such as [ITEM] mark positions for recommended items; in recommendation, candidates are scored and ranked.}
    \label{fig:model}
\end{figure*}

\begin{equation}
\mathbf{h}_n^{(l+1)} = \varphi \left( \sum_{r \in R} \sum_{j \in \mathcal{{N}}_r(n)} \frac{1}{c_{n,r}} \mathbf{W}_r^{(l)} \mathbf{h}_j^{(l)} + \mathbf{W}_0^{(l)} \mathbf{h}_n^{(l)} \right)
\end{equation}
where $\mathbf{h}_n^{(l)}$ denotes the embedding of node $n$ at layer $l$, $\mathcal{N}_r(n)$ is the set of neighbors of $n$ under relation $r$, $\mathbf{W}_r^{(l)}$ and $\mathbf{W}_0^{(l)}$ are the learnable weight matrices for relation $r$ and the self‐loop, respectively, $c_{n,r}$ is a normalization constant, and $\varphi$ is a nonlinear activation function. The resulting node embeddings form the final item embedding matrix $\mathbf{H}=[\mathbf{h}_\textit{i}^1, \mathbf{h}_\textit{i}^2, \dots, \mathbf{h}_\textit{i}^{n_{\textbf{H}}}]$.

\subsection{F-Former Module}

While R-GCN embeddings excel at encoding the rich relational topology of knowledge graphs, they inhabit a structural vector space misaligned with the purely linguistic semantics of pre-trained language models. To reconcile these two modalities, we introduce F-Former: a transformer-based alignment module, adapted from BLIP2’s Q-Former~\cite{li2023blip}, that projects graph-derived features into the PLM’s semantic space. F-Former is trained with contrastive, triplet and query-label matching objectives under a curriculum schedule, yielding unified representations that integrate graph knowledge and conversational context for more accurate recommendations.

\subsubsection{Information Encoding.}

We encode dialogue text and KG entities in parallel, then fuse them via cross‐modal queries. Specifically, dialogue tokens are embedded by a frozen RoBERTa~\cite{liu2019roberta}, producing a text embedding matrix that preserves lexical context. In parallel, RGCN generates structured entity embeddings.

To align these modalities, F-Former replaces the raw RGCN outputs with a fixed bank of \(K\) learnable query vectors $\mathbf{Q}_0 = \{\mathbf{q}_1,\dots,\mathbf{q}_K\}$, initialized from a normal distribution consistent with the encoder’s parameters. By feeding both queries and text through the same RoBERTa backbone, we minimize semantic distortion during projection.

Alignment proceeds via cross‐modal attention: each query attends first to the entity embeddings, extracting relational structure, and then to the text embeddings, capturing dialogue semantics. The attention mechanism is defined as:
\begin{equation}
\text{Attention}(\mathbf{Q}, \mathbf{H}) = \text{Softmax} \left( \frac{\mathbf{Q}\mathbf{W}_q (\mathbf{H}\mathbf{W}_k)^T}{\sqrt{D}}\right) \cdot \mathbf{H}\mathbf{W}_v
\end{equation}
where $\mathbf{W}_q$, $\mathbf{W}_k$, and $\mathbf{W}_v$ denote the query, key, and value transformation weights, respectively. $D$ denotes the hidden dimension, and the superscript $T$ in the equations denotes the transpose operation. The updated query vectors are generated:
\begin{equation}
\mathbf{Q}^{l+1}_{e} = \mathbf{Q}^{l}_{e} + \text{Attention}(\mathbf{Q}^{l}_{e}, \mathbf{H})\quad \mathbf{Q}^{0}_{e} =\mathbf{Q}_0
\end{equation}

After updating the query, we denote the final output as $\mathbf{Q}_{e}$. Then we use the RoBERTa model to generate context embedding vectors and use the [CLS] token $\mathbf{t}_{cls} \in  \mathbb{R}^{D}$ as the global representation of the entire context embedding. Analogous to the above cross‐attention operation, we obtain a query that fuses the text representations :
\begin{equation}
\mathbf{Q}^{l+1}_{t} = \mathbf{Q}^{l}_{t} + \text{Attention}(\mathbf{Q}^{l}_{t}, \mathbf{t}_{cls})\quad \mathbf{Q}^{0}_{t} =\mathbf{Q}_{e}
\end{equation}
Once $\mathbf{Q}_{e}\in\mathbb{R}^{K \times D}$ and $\mathbf{Q}_{t}\in\mathbb{R}^{K \times D}$ have been obtained, we compute their element-wise mean to derive the final fused representation $\mathbf{Q}$.

\subsubsection{Learning Tasks.}To enable better integration of entities and the context, we have designed three sub-learning tasks to assist the F-Former module in better learning how to perform cross-modal information fusion.

(1) \textit{Task 1: Batch-hard cross-modal contrastive learning:} To further enhance semantic alignment, we employ contrastive learning to optimize the model. By computing similarity between query-text pairs, we design a contrastive loss to ensure that semantically related queries and texts have higher similarity. We first calculate the similarity scores between normalized query $\mathbf{Q}\in\mathbb{R}^{B\times K \times D}$ and text embeddings $\mathbf{T}\in\mathbb{T}^{B \times D}$:
\begin{equation}
s_{q\to t}^{ijk}
= \frac{\mathbf{Q}_{i,k}\cdot \mathbf{T}_j}{\tau}
\quad
s_{t\to q}^{ijk}
= \frac{\mathbf{T}_i\cdot \mathbf{Q}_{j,k}}{\tau}
\end{equation}

where $\tau$ is a temperature coefficient, \(i\) is the index of the query sample within the batch, ranging from \(1\) to \(B\); \(j\) is the index of the text sample within the batch, also ranging from \(1\) to \(B\); and \(k\) is the index of the query slot for $\mathbf{Q}$, ranging from \(1\) to \(K\), $B$ is the batch size. $s_{q\to t}^{ijk}$ and $s_{t\to q}^{ijk}$ respectively represent the similarity scores from query to text and from text to query. 

Subsequently, we obtain the final similarity score through a max pooling function:
\begin{equation}
s_{q\to t}^{ij}
= \max_{k} s_{q\to t}^{ijk}
\quad
s_{t\to q}^{ij}
= \max_{k} s_{t\to q}^{ijk}
\end{equation}
For each anchor index $i$, define the positive scores and hardest negatives as:
\begin{equation}
p^i_{q\to t} = s_{q\to t}^{ii} 
\quad
h^i_{q\to t} = \max_{j\neq i} s_{q\to t}^{ij}
\end{equation}
\begin{equation}
p^i_{t\to q} = s_{t\to q}^{ii}
\quad
h^i_{t\to q} = \max_{j\neq i} s_{t\to q}^{ij}
\end{equation}
The vector $\mathbf{p}_{q\to t}$ comprises the diagonal entries of the similarity matrix \(\mathbf{S}_{q\to t}\), reflecting the similarity between each query slot and its corresponding text sample. Conversely, $\mathbf{h}_{q\to t}$ captures, for each query slot, the maximum similarity with any nonmatching text sample, that is, the hardest negative example in the query-to-text direction. Similarly, $\mathbf{p}_{t\to q}$ contains the diagonal entries of \(\mathbf{S}_{t\to q}\), measuring the similarity between each text sample and its matching query slot in the text-to-query direction, while $\mathbf{h}_{t\to q}$ records, for each text sample, the maximum similarity with any nonmatching query slot, highlighting the most challenging negative samples in the reverse direction.

The loss of bidirectional cross-entropy with smoothed labels is:
\begin{equation}
\mathcal{L}_{\mathrm{CE}}
= \frac12\Bigl[\mathrm{CE}\bigl(\mathbf{S}_{q\to t},\,\mathbf{y}\bigr)
+ \mathrm{CE}\bigl(\mathbf{S}_{t\to q},\,\mathbf{y}\bigr)\Bigr]
\end{equation}
where $\mathrm{CE}(\cdot,\cdot)$ denotes the cross‐entropy function and $\mathbf{y}$ is the ground‐truth label vector. 

Although minimizing cross-entropy loss encourages positive pairs to move closer and negative pairs to repel each other on average, it treats all negatives equally and thus may overlook the hardest impostors that lie near the decision boundary. To remedy this, we augment our training objective with a batch‐hard margin loss, which explicitly targets the most challenging negative example in each mini‐batch and sharpens the model’s discriminative capability:
\begin{equation}
\begin{split}
\mathcal{L}_{\mathrm{margin}} = \frac{1}{2B}\sum_{i=1}^B\Bigl[&\max\bigl(0,m + h^i_{\,q\to t}-p^i_{\,q\to t}\bigr) \\
& +\max\bigl(0,m + h^i_{\,t\to q}-p^i_{\,t\to q}\bigr)\Bigr]
\end{split}
\end{equation}
where $m$ is the margin, this hyperparameter controls the minimum difference between the similarity score of a positive sample and that of its hardest negative sample.

(2) \textit{Task 2: Recommendation feature triplet alignment:} To align the fused query representation with downstream recommendation features, we adopt a triplet‐margin loss that requires each true query–label pair to be closer by at least a margin \(m\) than the hardest negative in the batch. Given a batch of query‐slot features \(\mathbf{Q}\in\mathbb{R}^{B\times K \times D}\) and label embeddings \(\mathbf{R}\in\mathbb{R}^{B \times D}\), we first average the \(K\) slot vectors for each query to form a single \(D\)-dimensional fusion vector \(\mathbf{e}_i\). We then normalize both \(\mathbf{e}_i\) and its corresponding label embedding \(\mathbf{r}_i\), producing \(\tilde{\mathbf{e}}_i\) and \(\tilde{\mathbf{r}}_i\). Finally, we build a similarity matrix by computing the dot product between every \(\tilde{\mathbf{e}}_i\) and \(\tilde{\mathbf{r}}_j\), generating scores \(s_{ij}\) that serve as inputs to the tripletmargin objective.  

After obtaining the similarity matrix, operations similar to Formula (7) are adopted to obtain the positive score and the hardest negative for each batch index $i$:
\begin{equation}
p_i = s_{ii}
\qquad
n_i = \max_{j \neq i} s_{ij}
\end{equation}
\(s_{ij}\) measures the similarity between the \(i\)-th fused entity query and the \(j\)-th recommendation feature in the batch. Its diagonal entries \(s_{ii}\) correspond to each entity’s similarity with its own positive recommendation feature. 

Finally, the resulting triplet loss is:
\begin{equation}
\mathcal{L}_{\mathrm{triplet}}
= \frac{1}{B}\sum_{i=1}^B 
  \max\bigl(0,\;n_i - p_i + m\bigr)
\end{equation}
The triplet loss encourages each true entity–recommendation pair to lie at least $m$ closer than its most confounding negative.

(3) \textit{Task 3: Auxiliary query-label matching:} To further sharpen the alignment between the fused query slots and the embeddings of the downstream recommendation, we introduce an objective of auxiliary entity-text matching. Let $\tilde{\mathbf{e}}_i$ be the query representation of the fused entity as defined in Task 2, and let \(\tilde{\mathbf{r}}_i\) denote the corresponding normalized embedding of the recommendation. We employ a cosine‐embedding loss to encourage each $\tilde{\mathbf{e}}_i$ and \(\tilde{\mathbf{r}}_i\) pair to have high cosine similarity:
\begin{equation}
\mathcal{L}_{\mathrm{aux}}
= 
\frac{1}{B}\sum_{i=1}^B
\bigl(1 - \cos(\tilde{\mathbf{e}}_i,\,\tilde{\mathbf{r}}_i)\bigr)\,
\end{equation}
where \(\cos(\mathbf{u},\mathbf{v})\) measures cosine similarity.

\subsubsection{Curriculum Learning.} To ensure stable training, guide the model from coarse-grained targets to fine-grained targets, we adopt a three‐stage curriculum schedule that gradually introduces the three learning tasks. We specify $E_n$ to represent the number of epochs.

\paragraph{Stage I: Contrastive Warm‐Up.}  
For the first \(E_1\) epochs, we optimize only the batch‐hard contrastive loss:
\begin{equation}
\mathcal{L}_{s1} \;=\; \mathcal{L}_{\mathrm{CE}} + \mathcal{L}_{\mathrm{margin}}
\end{equation}
This encourages the queries to align at a coarse semantic level with their matching text embeddings before any harder negatives or auxiliary objectives are introduced.

\paragraph{Stage II: Triplet Refinement.}  
During the next \(E_n - E_1\) epochs, we add the triplet‐margin objective with a linearly ramped weight:
\begin{equation}
w_{\mathrm{triplet}}(e)
= 
\begin{cases}
0, & e < E_1,\\
\dfrac{e - E_1}{E_n - E_1}, & E_1 \le e < E_n
\end{cases}
\end{equation}
and optimize
\begin{equation}
\mathcal{L}_{s2}
= \mathcal{L}_{s1}
+ w_{\mathrm{triplet}}(e)\,\mathcal{L}_{\mathrm{triplet}}
\end{equation}
This stage sharpens the model’s ability to discriminate the hardest impostors while preserving the coarse alignment.

\paragraph{Stage III: Auxiliary Matching Consolidation.}  
In the final \(E_n - E_2\) epochs, we introduce the auxiliary query–label matching loss with a linear schedule:
\begin{equation}
w_{\mathrm{aux}}(e)
= 
\begin{cases}
0, & e < E_2,\\
\dfrac{e - E_2}{E_n-E_2}, & E_2 \le e < E_n
\end{cases}
\end{equation}
and optimize
\begin{equation}
\mathcal{L}_{cl}
= \mathcal{L}_{s2}
+ w_{\mathrm{aux}}(e)\,\mathcal{L}_{\mathrm{aux}}.
\end{equation}

By the end of curriculum training, the model has first learned coarse cross‐modal alignment, then fine‐grained discrimination via triplet alignment, and finally a tight proximity between query and label embeddings.

\subsection{Prompt Learning for Conversation Generation}

Our model builds on a pre‐trained language model (PLM) and adopts the UniCRS prompting strategy~\cite{wang2022towards} to handle recommendation and dialogue jointly. The F-Former fuses context and item embeddings: $\mathbf{H}_{\text{context-item}}\in\mathbb{R}^{B\times D}$. We then introduce a learnable conversation prefix $\mathbf{E}_{\text{conv}}$ and refine it via a two‐layer MLP \(\sigma\):
\begin{equation}
\mathbf{P}_{\text{conv}}
=\sigma\bigl(\mathbf{W}_c\,\mathbf{E}_{\text{conv}}+\mathbf{b}\bigr)
+\mathbf{E}_{\text{conv}}
\end{equation}
where \(\mathbf{W}_c\in\mathbb{R}^{D\times D}\) and \(\mathbf{b}\in\mathbb{R}^D\) are trainable matrix, and \(\sigma\) is a two-layer MLP with non-linear activations. We concatenate this with the fused embeddings to form the final prompt:
\begin{equation}
\mathbf{P}_{\text{conv}}
=\bigl[\mathbf{P}_{\text{conv}};\,\mathbf{H}_{\text{context-item}}\bigr]
\end{equation}
where $\left[;\right]$ denotes vector concatenation.

During training, the dialogue module minimizes the cross‐entropy loss:
\begin{equation}
\mathcal{L}_{\text{conv}}
=-\sum_{t=1}^{T}\log P\bigl(y_t \mid y_1, \dots, y_{t-1}, \mathbf{P}_{\text{conv}})
\end{equation}
where $y_t$ represents the $t$-th word in the target response, $T$ is the total length of the generated response, and $P \left( y_t \mid y_1, \dots, y_{t-1}, \mathbf{P}_{\text{conv}} \right)$ denotes the probability of generating the next word $y_t$ given the previously generated words $y_1, \dots, y_{t-1}$ and the prompt $\mathbf{P}_{\text{conv}}$.

To link dialogue and recommendation more closely, we follow UniCRS in re‐using generated responses to inform item prediction. We add a special token \texttt{[ITEM]} to the PLM vocabulary \(V\) and mask all item names in responses as \texttt{[ITEM]}. At inference, whenever the model emits \texttt{[ITEM]}, we post‐process by replacing it with the actual recommended item name.

\subsection{Prompt Learning for Item Recommendation}

The recommendation subtask aims to predict items that the user may find interesting by enriching prompt semantics with user–item interactions~\cite{gao2020making, xu2024sequence}. We also introduce a learnable recommendation prefix \(\mathbf{E}_{\text{rec}}\) and refine it via \(\sigma\):
\begin{equation}
\mathbf{P}_{\text{rec}}
= \sigma\bigl(\mathbf{W}_{r}\,\mathbf{E}_{\text{rec}} + \mathbf{b}\bigr)
+ \mathbf{E}_{\text{rec}}
\end{equation}
where \(\mathbf{W}_{r}\in\mathbb{R}^{D\times D}\) and \(\mathbf{b}\in\mathbb{R}^{D}\) are trainable.

\begin{table}[ht]
\centering
\caption{Statistics of the ReDial and INSPIRED datasets.}
\label{tab:dataset_stats}
\begin{tabular}{lcc}
\toprule
& \textbf{ReDial} & \textbf{INSPIRED} \\
\midrule
\# of conversations & 10,006 & 1,001 \\
\# of utterances & 182,150 & 35,811 \\
\# of words per utterance & 14.5 & 19.0 \\
\# of entities/items & 64,364/6,924 & 17,321/1,123 \\
\# of users & 956 & 1,482 \\
\bottomrule
\end{tabular}
\end{table}

To avoid overwhelming the original fused embeddings \(\mathbf{H}_{\text{context-item}}\), we apply a secondary fusion with scaling factor \(\lambda\):
\begin{equation}
\mathbf{H}'_{\text{context-item}}
= \mathbf{H}_{\text{context-item}}
+ \lambda\,\mathbf{H}
\end{equation}

We then concatenate the refined prefix, adjusted fusion, and a response template \(S\):
\begin{equation}
\mathbf{P}_{\text{rec}}
= \bigl[\mathbf{P}_{\text{rec}};\,\mathbf{H}'_{\text{context-item}};\,S\bigr]
\end{equation}

Given \(\mathbf{P}_{\text{rec}}\), the model minimizes the binary‐cross‐entropy recommendation loss:
\begin{equation}
\mathcal{L}_{\mathrm{rec}}
= -\sum_{n=1}^{N} \sum_{m=1}^{M}
\left[
  y_{n,m} \log\!\left( {Pr}_{n}(m) \right)
  + \left(1 - y_{n,m}\right)
    \log\!\left( 1 - {Pr}_{n}(m) \right)
\right]
\end{equation}
where \(N\) is the number of pairs of context-items, \(M\) the vocabulary size of the item, \(y_{n,m}\in\{0,1\}\) the ground truth label and \({Pr}_{n}(m)\) the softmax probability for the item \(m\).

Finally, we combine this with the curriculum learning loss \(\mathcal{L}_{cl}\):
\begin{equation}
\mathcal{L}'_{\text{rec}}
= \mathcal{L}_{\text{rec}}
+ \alpha\,\mathcal{L}_{cl}
\end{equation}
where \(\alpha\) balances curriculum learning’s contribution. The loss of conversation task \(\mathcal{L}_\text{conv}'\) follows a similar formula.

\section{Experiments}
We evaluated STEP in two public datasets using separate dialogue and recommendation metrics and performed module-wise ablation studies to validate the effectiveness of each proposed enhancement.

\subsection{Experimental Setup}

\textbf{Dataset:} To evaluate the performance of our model, we conducted experiments on the ReDial~\cite{li2018towards} and INSPIRED~\cite{hayati2020inspired} datasets. The ReDial dataset is an English conversational recommendation dataset focused on movie recommendations, created by crowdsourced workers on Amazon Mechanical Turk (AMT). Similar to ReDial, the INSPIRED dataset is also an English conversational recommendation dataset for movies, but it is smaller in scale. These two datasets are widely used for evaluating CRS models. The statistics of the two datasets are presented in Table~\ref{tab:dataset_stats}.

\textbf{Baselines:}
For baselines, we compare against two PLM‐based dialogue generators—\textit{DialoGPT}~\cite{DBLP:conf/acl/ZhangSGCBGGLD20} and OpenAI’s \textit{GPT‐3.5‐turbo} and \textit{GPT‐4}~\cite{achiam2023gpt}—and six representative CRS approaches: \textit{ReDial}~\cite{li2018towards} and \textit{KBRD}~\cite{chen2019towards} as early auto‐encoder and KG‐augmented methods; \textit{KGSF}~\cite{zhou2020improving} and \textit{UniCRS}~\cite{wang2022towards} as knowledge‐enhanced and prompt‐tuning frameworks; and \textit{VRICR}~\cite{zhang2023variational} and \textit{DCRS}~\cite{dao2024broadening}, which utilize variational Bayesian pre‐training and retrieval‐augmented conversational understanding, respectively.

\begin{table}[ht]
  \centering
  \small
  \setlength\tabcolsep{4pt} 
  \caption{Results of the recommendation task. Results marked with * show noticeably larger improvements over the best baseline (t-test, $p$-value $<$ 0.05).}
  \label{tab:recommendation_results}
  \begin{tabularx}{\columnwidth}{
      @{}l
      *{3}{>{\centering\arraybackslash}X}
      |
      *{3}{>{\centering\arraybackslash}X}
      @{}
    }
    \toprule
    \textbf{Datasets}
      & \multicolumn{3}{c}{\textbf{ReDial}}
      & \multicolumn{3}{c}{\textbf{INSPIRED}} \\
    \cmidrule(lr){2-4}\cmidrule(lr){5-7}
    \textbf{Models}
      & \textbf{R@1} & \textbf{R@10} & \textbf{R@50}
      & \textbf{R@1} & \textbf{R@10} & \textbf{R@50} \\
    \midrule
    ReDial         & 0.023 & 0.129 & 0.287 & 0.003 & 0.117 & 0.285 \\
    DialoGPT       & 0.030 & 0.173 & 0.361 & 0.024 & 0.125 & 0.247 \\
    KBRD           & 0.033 & 0.175 & 0.343 & 0.058 & 0.146 & 0.207 \\
    KGSF           & 0.036 & 0.177 & 0.363 & 0.058 & 0.165 & 0.256 \\
    GPT-3.5-turbo  & 0.039 & 0.168 & --    & 0.051 & 0.150 & --    \\
    GPT-4          & 0.045 & 0.194 & --    & 0.091 & 0.194 & --    \\
    VRICR          & 0.057 & 0.251 & 0.416 & 0.056 & 0.179 & 0.345 \\
    UniCRS         & 0.051 & 0.224 & 0.428 & 0.090 & 0.277 & 0.426 \\
    DCRS           & 0.070 & 0.248 & 0.434 & 0.093 & 0.226 & 0.414 \\
    \midrule
    \textbf{STEP}  & \textbf{0.081*} & \textbf{0.256} & \textbf{0.440}
                   & \textbf{0.131*} & \textbf{0.301*} & \textbf{0.433} \\
    \bottomrule
  \end{tabularx}
\end{table}
\begin{table}[ht]
\centering
\caption{Ablation Study on Recommendation Task (ReDial)}
\label{tab:ablation_recall}
\scalebox{1}{%
\begin{tabular}{lccc}
\toprule
\textbf{Model} & \textbf{Recall@1} & \textbf{Recall@10} & \textbf{Recall@50} \\
\midrule
- \textit{w}/\textit{o} \textit{CL} & 0.071 & 0.238 & 0.417 \\
- \textit{w}/\textit{o} \textit{Task1} & 0.070 & 0.234 & 0.411 \\
- \textit{w}/\textit{o} \textit{Task2} & 0.074 & 0.241 & 0.422 \\
- \textit{w}/\textit{o} \textit{Task3} & 0.076 & 0.242 & 0.435 \\
\midrule
\textbf{STEP} & \textbf{0.081} & \textbf{0.256} & \textbf{0.440} \\
\bottomrule
\end{tabular}
}
\end{table}

\textbf{Evaluation Metrics:} Following previous CRS work~\cite{wang2022towards,zhang2023variational}, we use different metrics to evaluate the recommendation and conversation tasks separately. For the recommendation task, we adopt Recall@k (k=1, 10, 50) to measure the fraction of items of ground truth successfully recovered within the recommended list k top. For the conversation task, we use Distinct-n (n=2, 3, 4) at the word level to assess the diversity of generated responses.

\textbf{Implementation Details:} All experiments were conducted on a single NVIDIA L20 GPU (48\,GB). We build on DialoGPT-small with a frozen RoBERTa-base encoder and a single R-GCN layer following DCRS. After tuning, we set the soft prefix length to 16 (ReDial) / 8 (INSPIRED), the query length \(K=32\), and the curriculum epochs \(E_1=2\), \(E_2=3\), and \(E_n=5\). Optimization uses AdamW~\cite{DBLP:conf/iclr/LoshchilovH19} with batch sizes of 54 (recommendation) and 24 (conversation), a pretrain LR of \(5\times10^{-4}\) and finetune LR of \(1\times10^{-4}\), balancing losses with \(\alpha=0.5\). Zero-shot LLM baselines (GPT-3.5-turbo, GPT-4) follow He et al.~\cite{he2023large}, while other baselines use the CRSLab toolkit~\cite{DBLP:conf/acl/ZhouWZSCZLW21}.

\subsection{Evaluation on Recommendation Task}

In this section, we evaluate the effectiveness of our model on the recommendation task through various experiments.

\textbf{Automatic Evaluation:} 
Table~\ref{tab:recommendation_results} compares various methods on the recommendation task. Among the CRS approaches, DCRS excels in using knowledge-aware contrastive learning to retrieve and learn from example dialogues, thus enriching the prompts. Its use of ``contextual'' and ``knowledge-enhanced'' prompts bridges the gap between generation and recommendation. UniCRS further validates the effectiveness of PLMs in a unified conversational recommendation framework through cross-modal knowledge fusion. KG-based methods also perform strongly, underscoring the value of knowledge graphs in capturing user interests and enriching conversation semantics.

In particular, PLM-based methods, based solely on language modeling, achieve results comparable to KBRD, highlighting the advantages of contextual understanding. Similarly, LLM-based methods, such as GPT-3.5-turbo and GPT-4, demonstrate impressive performance due to their superior language generation capabilities and advanced contextual reasoning. However, both PLM-based and LLM-based approaches struggle to effectively leverage context to locate and retrieve corresponding external knowledge for information enrichment.

Our proposed STEP outperforms all baselines. This remarkable improvement over strong baselines DCRS \& UniCRS can be observed in terms of Recall@1 (+15.7\% for ReDial, +40.8\% for INSPIRED), Recall@10 (+3.2\% for ReDial, +8.6\% for INSPIRED), and Recall@50 (+1.4\% for ReDial, +1.6\% for INSPIRED).
In general, STEP effectively fuses knowledge graphs and contextual prompts to guide PLM generation, seamlessly integrating relevant KG information with dialogue context. This prompt-based strategy not only increases flexibility and adaptability but also leads to better recommendation performance.

\textbf{Ablation Study:} Our model is designed with a series of prompt components to enhance the performance of CRS. To verify the effectiveness of each component, we conducted ablation experiments on the ReDial dataset and reported the results for Recall@1, Recall@10, and Recall@50. We sequentially considered the removal of the curriculum learning (\textit{w}/\textit{o} \textit{CL}), batch-hard cross-modal contrastive learning (\textit{w}/\textit{o} \textit{Task1}), recommendation feature triplet alignment (\textit{w}/\textit{o} \textit{Task2}) and auxiliary query-label matching (\textit{w}/\textit{o} \textit{Task3}). The results are shown in Table~\ref{tab:ablation_recall}. 

As seen, each learning component contributes a unique yet complementary effect on model performance. Removing curriculum learning forfeits the gradual “easy-to-hard” progression, which undermines the model’s ability to establish a robust foundation for semantic alignment in the early stages. Omitting batch-hard contrastive learning substantially weakens the model’s capacity to discriminate between highly similar instances, impairing its ability to capture fine-grained distinctions between queries and texts. Eliminating the triplet alignment objective prevents the downstream recommendation module from further reinforcing the mapping between query representations and label embeddings, thereby reducing overall recommendation effectiveness. Finally, discarding the auxiliary matching loss removes the critical fine-tuning step that refines the proximity between fused query embeddings and target labels, resulting in suboptimal alignment at a detailed level. These results indicate that curriculum learning, contrastive warm-up, triplet refinement, and auxiliary matching each address different facets of the training process and together form a coherent coarse-to-fine curriculum that enhances recommendation performance.

\subsection{Evaluation on Conversation Task}

In this section, we evaluate the effectiveness of our model on the conversation task through various experiments.

\begin{table}[ht]
  \centering
  \small
  \setlength\tabcolsep{4pt} 
  \caption{Automatic evaluation results on the conversation task. Results marked with * show noticeably larger improvements over the best baseline (t-test with $p$-value $<$ 0.05).}
  \label{tab:conversation_results}
  \begin{tabularx}{\columnwidth}{
      @{}l
      *{3}{>{\centering\arraybackslash}X}
      |
      *{3}{>{\centering\arraybackslash}X}
      @{}
    }
    \toprule
    \textbf{Dataset}
      & \multicolumn{3}{c}{\textbf{ReDial}}
      & \multicolumn{3}{c}{\textbf{INSPIRED}} \\
    \cmidrule(lr){2-4}\cmidrule(lr){5-7}
    \textbf{Models}
      & \textbf{Dist-2} & \textbf{Dist-3} & \textbf{Dist-4}
      & \textbf{Dist-2} & \textbf{Dist-3} & \textbf{Dist-4} \\
    \midrule
    ReDial   & 0.058 & 0.204 & 0.442 & 0.359 & 1.043 & 1.760 \\
    KBRD     & 0.085 & 0.163 & 0.252 & 0.416 & 1.375 & 2.320 \\
    KGSF     & 0.114 & 0.204 & 0.282 & 0.583 & 1.593 & 2.670 \\
    DialoGPT & 0.286 & 0.352 & 0.291 & 1.995 & 2.633 & 3.237 \\
    VRICR    & 0.233 & 0.292 & 0.482 & 0.853 & 1.801 & 2.827 \\
    UniCRS   & 0.404 & 0.518 & 0.832 & 3.039 & 4.657 & 5.635 \\
    DCRS     & 0.608 & 0.905 & 1.075 & 3.950 & 5.729 & 6.233 \\
    \midrule
    \textbf{STEP}
      & \textbf{0.637} & \textbf{1.017*} & \textbf{1.294*}
      & \textbf{3.968}  & \textbf{5.856} & \textbf{6.631*} \\
    \bottomrule
  \end{tabularx}
\end{table}

\begin{table}[ht]
\centering
\caption{Ablation Study on Conversation Task (ReDial)}
\label{tab:ablation_distinct}
\scalebox{1}{%
\begin{tabular}{lccc}
\toprule
\textbf{Model} & \textbf{Distinct@2} & \textbf{Distinct@3} & \textbf{Distinct@4} \\
\midrule
- \textit{w}/\textit{o} \textit{CL} & 0.536 & 0.840 & 1.058 \\
- \textit{w}/\textit{o} \textit{Task1} & 0.489 & 0.773 & 0.850 \\
- \textit{w}/\textit{o} \textit{Task2} & 0.583 & 0.872 & 1.075 \\
- \textit{w}/\textit{o} \textit{Task3} & 0.605 & 0.901 & 1.113 \\
\midrule
\textbf{STEP} & \textbf{0.637} & \textbf{1.017} & \textbf{1.294} \\
\bottomrule
\end{tabular}
}
\end{table}

\textbf{Automatic Evaluation:}
Table~\ref{tab:conversation_results} shows the results of the automatic evaluation for the generation of conversations. STEP achieves the highest performance, especially on Distinct-n (n=2, 3, 4), suggesting improved diversity and richness in generated dialogues. While KG-based methods (e.g., KBRD, KGSF, VRICR) leverage external knowledge to enhance dialogue understanding, STEP integrates enhanced prompt design and the F-Former module for deeper semantic alignment between KG and dialogue context, enabling more targeted and diverse responses.

Compared to UniCRS and DCRS, STEP delivers greater diversity and informativeness through more effective knowledge fusion. Although UniCRS employs a unified architecture and knowledge-enhanced prompts, it still lags behind STEP in semantic depth and diversity. DCRS uses knowledge-aware contrastive learning to augment prompts with relevant example dialogues, but it can be limited in handling complex scenarios. In contrast, STEP captures deeper contextual nuances to generate more coherent and contextually aligned responses.

\textbf{Ablation Study:} The proposed prompt design significantly improves the performance on the conversation task. To verify the role of each component, we also conducted an ablation study on the ReDial dataset, using Distinct@2,3,4 as evaluation metrics. In the experiment, we sequentially removed curriculum learning (\textit{w}/\textit{o} \textit{CL}), batch-hard cross-modal contrastive learning (\textit{w}/\textit{o} \textit{Task1}), recommendation feature triplet alignment (\textit{w}/\textit{o} \textit{Task2}) and auxiliary query-label matching (\textit{w}/\textit{o} \textit{Task3}). The results are shown in Table~\ref{tab:ablation_distinct}. 

\begin{table}[ht]
\centering
\caption{One case extracted from the ReDial dataset.}
\label{tab:redial_case}
\normalsize
\scalebox{0.95}{%
\begin{tabular}{p{8cm}}
\toprule
\rowcolor{white} 
\textbf{Context} \\
\rowcolor{gray!20} Recommender: Hello there. Can I help you find a good movie? \\
\rowcolor{gray!20} User: I like dramas and old black and white movies, I've seen \textbf{\textit{Rear Window}}. \\
\midrule
\rowcolor{white}
\textbf{Response} \\
\rowcolor{pink!20} UniCRS: I would watch \textbf{\textit{Gone with the Wind}}.\\
\rowcolor{orange!20} DCRS: I know of that one. How about \textbf{\textit{Casablanca}}? \\ 
\rowcolor{cyan!20} \textcolor{red}{STEP}: Have you seen \textbf{\textit{Vertigo}}? I am a big fan of it. \\ 
\bottomrule
\end{tabular}
}
\end{table}

The results of the ablation of the conversation task reveal how each training component shapes the model’s ability to generate diverse lexical responses. Omitting curriculum learning removes the gradual introduction of harder objectives, which in turn constrains the model’s capacity to explore varied linguistic patterns and yields more repetitive n-grams. Eliminating the batch-hard contrastive objective has the most pronounced effect on diversity: without this fine-grained discrimination, the model struggles to distinguish subtly different contexts and resorts to common or boilerplate phrases. In contrast, removing the triplet alignment stage only slightly diminishes the distinct n scores, indicating that its main benefit lies in the consistency of the downstream recommendation rather than the variability of the conversation. Finally, abrogating the auxiliary query-label matching loss produces a modest drop in diversity, suggesting that this fine-tuning step, while secondary, still contributes to weaving in novel lexical choices.  Collectively, these findings underscore that curriculum scheduling and contrastive warm-up are critical for fostering conversational richness, whereas later alignment tasks play supportive roles in refining response novelty.


\subsection{Hyper-parameters Optimizing}
Preliminary experiments suggest that prefix length and query count have little effect on Distinct@k, so in this section we focus on hyperparameter tuning of the more important recommendation task.

\textbf{Prefix Length Analysis:} In our study, we systematically evaluated prefix lengths of 4, 8, 16, and 24 tokens to assess their effect on Recall@1 and Recall@50. Our results show that the optimal prefix length varies by dataset: on ReDial, performance steadily increases and reaches its maximum at 16 tokens, whereas on INSPIRED, the highest recall is achieved with just 8 tokens. This clearly illustrates that the ideal context window must be tailored to the dialogue characteristics of each data set.

\begin{figure}[htbp]
\centerline{\includegraphics[width=1 \linewidth]{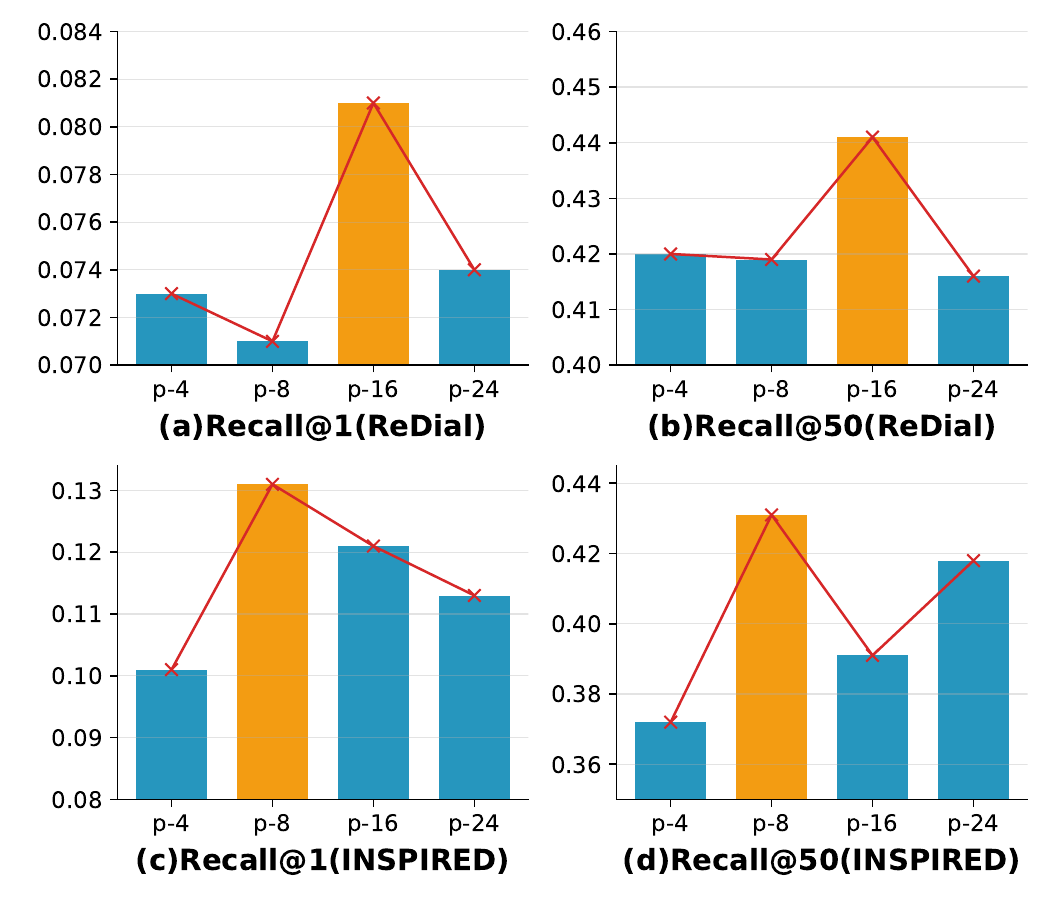}}
\caption{Hyper-parameters optimizing (prefix length) on item recommendation.}
\label{fig:prefix}
\end{figure}

Figures~\ref{fig:prefix} (a) and (c) depict the Recall@1 curves, while Figures~\ref{fig:prefix} (b) and (d) show Recall@50. In both metrics, ReDial’s recall improves with longer prefixes up to 16 tokens before declining, which indicates an information overload beyond this point, while INSPIRED peaks at 8 tokens and diminishes thereafter. We attribute this divergence to the distinct conversational styles: ReDial comprises extended, multi-turn exchanges rich in movie mentions and genre shifts, which require a broader context window to capture salient cues; INSPIRED, by contrast, follows a concise, structured Q\&A format focused on a single recommendation target, where additional context can introduce low-relevance content and dilute model focus.

We trace the discrepancy in optimal prefix lengths to four key factors: dialogue length, information density, interaction structure, and noise profile. ReDial’s longer, more complex dialogues, with high information density and abrupt topic transitions, require a wider context to maintain coherence and user intent across turns. Conversely, INSPIRED’s uniform utterance style and lower noise level allow effective recommendations with a shorter context. Together, these dataset-specific factors determine the prefix length that best balances contextual completeness with relevance.

\textbf{Query Length Analysis:} When analyzing query length, we explored the impact of different query numbers (24, 32, 40, 48) on Recall@1 and Recall@50. The results show that increasing the number of queries initially improves recall, reaching the best performance at 32 queries, after which recall begins to decline.

\begin{figure}[htbp]
\centerline{\includegraphics[width=1 \linewidth]{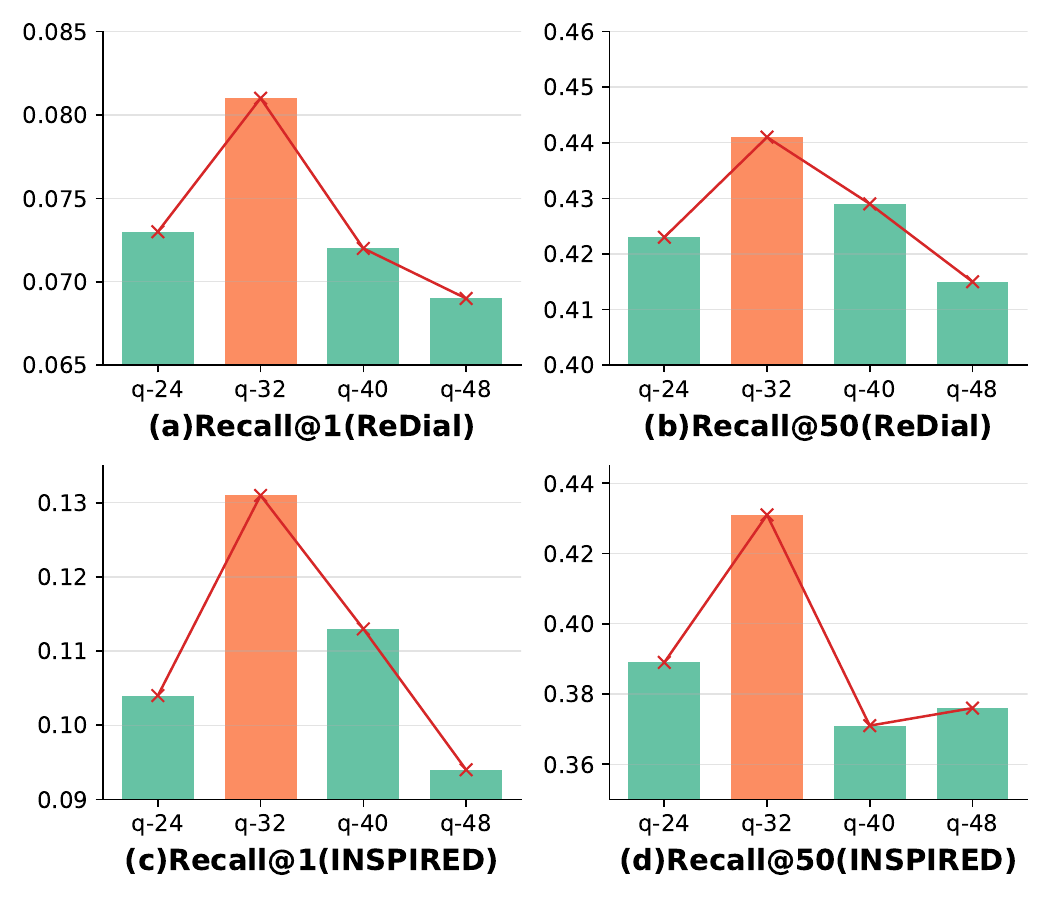}}
\caption{Hyper-parameters (query length) optimizing on item recommendation.}
\label{fig:query}
\end{figure}

For both Recall@1 (Figures~\ref{fig:query} (a) and (c)) and Recall@50 (Figures~\ref{fig:query} (b) and (d)), performance peaks at 32 queries. With fewer queries (e.g., 24), the model struggles to capture sufficient recommendation diversity, while increasing beyond 32 (e.g., 40 or 48) introduces redundancy and noise among query representations. This follows the law of diminishing returns: extra queries no longer yield meaningful new information but instead impede overall recall performance.

The optimal performance at 32 queries suggests that a moderate number of queries strikes a balance between capturing diverse information and maintaining model stability, enabling the model to provide accurate and relevant recommendations without suffering from excessive computational overhead or information noise.

\subsection{Case Study}

To evaluate STEP on conversational recommendation, we compared it against DCRS and UniCRS on the ReDial dataset, with a representative case shown in Table~\ref{tab:redial_case}. When asked for “some classic suspense movies in the style of \textit{Rear Window},” STEP replies in the third turn with \textit{Vertigo}, leveraging F-Former’s cross-attention to fuse KG relations and dialogue context, thus capturing both the “classic” attribute and the specific suspense focus. In contrast, DCRS suggests \textit{Casablanca} and UniCRS \textit{Gone with the Wind}—although both models correctly identify the broader “classic” dimension, they ignore the suspense-related attributes encoded in the knowledge graph, resulting in recommendations that miss the mark. This oversight of KG‐derived suspense cues highlights their inability to fully align dialogue context with the most pertinent genre information.

This example highlights STEP’s ability to extract user intent, dynamically filter and weight knowledge-graph signals, and synthesize semantic relationships across modalities. By integrating entity embeddings and conversational history, STEP delivers more relevant, diverse, and personalized recommendations. This dynamic alignment of graph and dialogue semantics enhances recommendation accuracy and user engagement in conversational settings.

\section{Conclusion}

STEP is a novel conversational recommendation system that integrates PLM and KG through curriculum learning and prompt learning. Its F-Former architecture effectively aligns KG information with dialogue context, enhancing recommendation accuracy. The experimental results demonstrate that STEP surpasses existing approaches in both the effectiveness of the recommendation and the quality of the dialogue. Future work will integrate multimodal inputs to further enhance recommendation effectiveness and user engagement.

\begin{acks}
This work was supported in part by BNSF(L233034, L253004), Fundamental Research Funds for the Beijing University of Posts and Telecommunications(No.\,2025TSQY01), Major Research Program of the Zhejiang Provincial Natural Science Foundation (LD24F020015), Guangdong Basic and Applied Basic Research Foundation (2025A15
15010739), Guangzhou Science and Technology Program (2024A04J6
317), NSFC (62306287) and Scientific Foundation for Youth Scholars of Shenzhen University (No.868-000001032902).
\end{acks}

\section{GenAI Usage Disclosure}

In the preparation of this paper, we used OpenAI’s ChatGPT-o4-mini-high model to polish the language and improve the readability of the text. No generative AI tools were employed in the conception, design, or execution of the research, including data collection, analysis, figure preparation, code development, or experimental procedures.


\bibliographystyle{ACM-Reference-Format}
\balance


\end{document}